# VLM@school – Evaluation of AI image understanding on German middle school knowledge

René Peinl[1] and Vincent Tischler[1]

[1]Hof University of Applied Sciences, Hof, Germany
`rene.peinl@hof-university.de`

**Abstract.** This paper introduces a novel benchmark dataset designed to evaluate the capabilities of Vision Language Models (VLMs) on tasks that combine visual reasoning with subject-specific background knowledge in German language. In contrast to widely used English-language benchmarks that often rely on artificially difficult or decontextualized problems, this dataset draws from real middle school curricula across nine domains including mathematics, history, and biology. The benchmark includes over 2,000 open-ended questions grounded in 486 images, ensuring that models must integrate visual interpretation with factual reasoning rather than rely on superficial textual cues. We evaluate thirteen state-of-the-art open-weight VLMs across multiple dimensions, including domain-specific accuracy and performance on adversarial crafted questions. Our findings reveal that even the strongest models achieve less than 45% overall accuracy, with particularly poor performance in music, mathematics, and adversarial settings. Furthermore, the results indicate significant discrepancies between success on popular benchmarks and real-world multimodal understanding. We conclude that middle school-level tasks offer a meaningful and underutilized avenue for stress-testing VLMs, especially in non-English contexts. The dataset and evaluation protocol serve as a rigorous testbed to better understand and improve the visual and linguistic reasoning capabilities of future AI systems.

## 1 Introduction

AI development is heavily driven by benchmarks. ImageNet [1] kicked off this development for image classification and object recognition that was continued in other domains e.g. with LibriSpeech [2] for automatic speech recognition and MMLU (Pro), the Massive Multitask Language Understanding benchmark [3]. However, there is increasing critique with that benchmark-driven approach [4].

Benchmarks have been criticized for overpromising their utility [5], being susceptible to manipulation [6], assessing inappropriate or irrelevant aspects of model performance [7], and lacking practical applicability in real-world scenarios [8]. They often suffer from insufficient documentation [9], attain credibility through arbitrary community consensus rather than rigorous validation [10], and reflect problematic cultural assumptions [11]. Additionally, benchmarks tend to be narrow in scope, predominantly focused on English [12], while adhering to a one-time testing paradigm [13] that neglects multimodal capabilities such as visual and auditory processing [14]. These kinds of benchmarks also tend to ignore the contextual and social dimensions of AI harm, which cannot be adequately captured in abstract, decontextualized evaluation settings devoid of human interaction [14].

For visual understanding, MMMU [15], the massive multi-discipline multimodal understanding and reasoning benchmark, is presumably one of the most cited benchmarks. It can be seen as a counterpart of the text-only MMLU for vision language models (VLMs) and is also an important part of the OpenVLM leaderboard on Huggingface [16], that tries to capture a broad range of VLM capabilities by running multiple different benchmarks. Despite this, it is again only available in English language and heavily relies on multiple choice questions (94%), which cannot completely capture the capabilities necessary for real world tasks.

Therefore, the **goal of this work** is to provide a comprehensive set of benchmark tasks for VLMs across several domains in German language, while at the same time avoiding common shortcomings of existing datasets.

In contrast to the current trend to create increasingly hard problems for AI that are also increasingly hard to solve for humans [17, 18], we hypothesize that already basic questions that target middle school level students (grades eight to ten in German school system) are challenging for current state-of-the-art open-weight VLMs, if they require both visual understanding, reasoning as well as background knowledge within the tested domains (e.g. biology, history, geography). This can be seen as a continuation of the work done with the FactSight dataset [19], which has already shown that even for purely factual data, performance in German is lower than in English for questions that don't require human experts to be solved correctly.

The remainder of the paper is structured as follows. First, pros and cons of existing benchmarks for VLMs are discussed in the related work section. Then, it is described how our dataset was constructed and how it is composed. We move on with discussing model selection and evaluation criteria in sections four and five. Results are presented in section six and analyzed in detail in section seven. Some limitations of our work are highlighted in section eight before ending with a conclusion and outlook.

## 2      Related Work

**MMMU** [15] is still the main reference for visual understanding and used by big players in the AI industry like Google [20], Meta [21] and Anthropic [22] to showcase the capabilities of their latest models, despite well-known limitations like the text-only solvability of some benchmark questions, limited options in multiple-choice formats, and its' failure to challenge the models' abilities to jointly understand different modalities in a rigorous way [23]. We therefore follow the approach of MMMU-Pro that addresses these shortcomings by (1) filtering questions answerable by text-only models, (2) including questions that have the question text as part of the image and (3) avoiding the limited option space by asking open-ended question [23]. This led to significant drops in the performance of leading VLMs of 16.8% to 26.9% for **MMMU-Pro** compared to MMMU. Current scores of the best VLMs on MMMU are between the lower bound of human experts (76%) and the upper bound (88%), e.g. Gemini 2.5 with 79% (both Pro and Flash) or OpenAI's o3 with 83%. When the paper was published, the best model achieved 59% across the disciplines art & design, business, science, health & medicine, humanities & social sciences as well as tech & engineering, which can be seen as a reference for our own dataset.

One challenging aspect of MMMU is the use of illustrations and charts that seem to be underrepresented in the training data of most VLMs compared to photos. Other benchmarks like **ChartQA** [24] have noticed that before and try to fill the gap, but only for a limited number of different types of images like line, bar and pie charts. Similar to our dataset, it also requires models to do basic calculations like adding two numbers, counting or calculating a difference. Current SOTA for ChartQA is at 90% (Llama 4).

The **AI2D** dataset [25] is also closely related to our endeavor and contains 5,000 diagrams of diverse types together with 15,000 multiple choice questions. It was already constructed in 2016 and therefore seems too simple for modern VLMs. Questions like "the diagram depicts the life cycle of a) frog b) bird c) insecticide d) insect" and the image contains both a picture of a frog as well as the word frog are not challenging. Some questions require reasoning like deriving from a food web that the mountain lion is eating deer and therefore a decrease in the lion population will lead to an increase in the deer population. However, a test with a modern LLM shows that even without the picture, the answer can be given correctly.

**MMVet** [26] is a rather small benchmark with only 200 images and 218 questions. It is however still popular, since it is both challenging and requires a combination of skills from the VLM like recognition, background knowledge, OCR, spatial awareness and math. It also asks open-ended questions like we do in our dataset. The current SOTA is over 80% on MMVet.

**SciVerse** [27] is a dataset from 2025 that systematically tests how textual context influences the VLMs ability to answer questions in chemistry, biology and physics correctly. To do that, they remove helpful words from the question text, so that the context has to be derived from the image instead. They call the different kinds of questions knowledge-rich, knowledge-lite and knowledge-free. They also test a version where the text of the knowledge-free question is put in the image instead (vision only). Results show that no additional knowledge is needed. The differences between the knowledge-rich and knowledge-free versions are minimal. For vision-only, the study finds a drop in accuracy of about 4-5% absolute for the better models. We therefore included some questions that are printed in the image in our dataset. For some questions (especially religion), we provided a little textual context (e.g. the hint "the image depicts a religious scene") because otherwise questions like "who is the person on the left of the image" are not answerable unambiguously.

Another recent dataset is called **MapBench** [30]. It compiles over 1,600 map space path-finding problems associated with 100 diverse maps images including indoor (e.g. museums) and outdoor facilities (e.g. Zoos). The evaluation is done with a score that compares the suggested path of the models with the shortest path, so a score of one would be a perfect result. They only test four different models, Llama 3.2 11B, Qwen 2 VL 7B, GPT-4o and GPT-4o mini, once with a standard prompt and once with a chain-of-thought prompt. Best models achieve a score between 1.2 and 1.8 on easier maps and slightly above 2 for harder maps. We also have maps in both geography as well as history in our datasets, but do not calculate paths. **Exams-V** [31] is probably the closet to our dataset and can be seen as a multi-lingual version of MMMU. It covers exams from 20 domains across eleven languages including 819 German questions, of which only 144 are visual questions. The rest of the questions are purely textual, although presented as an image. The distribution of domains across languages is not homogeneous. For German, only Business, Chemistry and Physics are included in the training set and only 14 questions belong to Chemistry. The German test set additionally includes tourism (11) and geography (5) questions, but the majority comes from Physics (185 out of 258). All questions are directly included in the image, not presented as additional text. The published dataset has no ground truth for the test set, so we could not compare results with our dataset.

**MMMBench** [32] is another dataset that consists of image text pairs in multiple languages. Besides English and Chinese, it also includes Turkish, Russian and Arabic. Li et al [33] translated a subset of ImageNet, VQA v2 and OK-VQA into 80 languages, including most European ones. **xGQA** [34] is a translation of GQA with over 9,000 questions into seven additional languages, including German. Their evaluation results show a gap of 20% and more between English and German language. Astonishingly, they only translated the questions, not the answers, which are still in English language. Most of these datasets rely on machine translation, although Park et al [35] find that translation artefacts can significantly affect the performance of models and can shift the meaning far enough to change the correct answer from yes to no. To avoid these kinds of problems, we only use questions created by German native speakers.

Summing up existing work, we found no benchmark with German questions and answers created by native speakers instead of machine translation. Our dataset fills this gap. Our evaluation also exceeds existing work by using free text answers instead of multiple choice. It takes up lessons from MMMU Pro and rigorously filters questions answerable by LLMs and includes question text in the image for some questions.

## 3    Dataset

We collect questions from German school textbooks from various grades (8-10) and school types (Gymnasium/grammar school, Realschule/secondary school). We also include images from non-school books that target pupils and intermediate practitioners of the field, e.g. arts and sports. Associated questions partly stem from the same books but are mostly self-constructed by the authors to ensure alignment with the design goals. Images in the categories music and religion were supplemented with pictures from the internet. Overall, the dataset comprises 486 images and 2038 questions across 9 domains (school subjects, see Table 1). Image sizes were limited to 1300px in both directions. The average size of an image is between 1.3M pixel (history) and 500k pixel (math). The dataset is publicly available[1].

For math questions we make sure that humans can calculate the numbers in their minds and do not need to do textual **calculations** to find the correct answer. This means avoiding large numbers or any numbers with lots of digits (e.g. 1.2573) but instead using easy to calculate numbers like numbers between 1 and 99 or multiples of ten or hundred (like 2000-700). Instead, we focused on questions that require background knowledge like sums of angles in a triangle or calculating the outline of a circle. Instead

---

[1] https://iisys-hof.github.io/vlms-at-school/

Evaluation of AI image understanding on German knowledge        5of asking for a pure number, we explicitly told the models to use Pi as a variable in the calculations to make them easier (e.g. 6Π+14 instead of 32.849 as the result). In another case, we told the model to use Π=3.1 to make the calculation easier. Our focus was not so much on the calculation, but the model's ability to derive the term from the picture. Pure math is already tested well in LLM benchmarks.

Table 1.   Number of images and questions per domain of the dataset

| Domain | images | questions | Example question (translated to English) |
|---|---|---|---|
| Arts | 57 | 286 | To which stylistic period of art does this work belong to? |
| Biology | 33 | 169 | What is the heart rate at the end of the test? |
| Geography | 46 | 251 | On which longitude was the Earth's magnetic pole according to the map in 1973? |
| History | 52 | 264 | According to the map, what is the oldest Aragonese legacy of Charles V? |
| Math | 112 | 308 | Calculate the area of figure A. |
| Music | 34 | 166 | Which note can be found in all chords depicted? |
| Physics | 52 | 234 | What is the amperage of light bulb X? |
| Religion | 65 | 206 | Which religious scene is depicted here? |
| Sports | 35 | 154 | What is the number of the player defending in the post? |

It is a well-known shortcoming of large language models (LLMs) and consequently VLMs to have problems stating that they don't know something, or that a question is unanswerable [36, 37]. We therefore also include questions that cannot be answered based on the given picture. We call these adversarial questions. They are phrased intentionally misleadingly and refer to concepts that are semantically close to those present in the image. We ask e.g. for the harp of an angel in a picture showing "the proclamation to Mary", which is not present in the picture, but a typical object often depicted together with angels.

## 4    Model Selection

To find out whether our dataset is both challenging and selective, we tested a number of open weight VLMs. Our first selection criteria were the models results in the open VLM leaderboard available on Huggingface, which uses VLMEvalKit [16] and a number of popular benchmarks to generate an aggregated number. This is to the authors knowledge the best available indicator of model capabilities across a broad range of disciplines and task types.

The second aspect to consider was model availability, so we filtered for open models. We didn't filter for model size, but naturally, bigger models rank better than smaller ones, although some small models outperform bigger ones. We excluded older models that were superseded by more recent models from the same family, so we only tested InternVL3 models and Qwen 2.5 VL, but not InternVL2.5 and Qwen 2 VL. This left us with a list with only three families of models, namely InternVL, Qwen and Ovis, which all use Qwen 2.5 as an LLM backbone and all stem from Chinese companies. Since we are evaluating models on German language, we decided to also include a few US and European models, especially those missing in the Open VLM leaderboard. This gave us the final list of thirteen competitors shown in Table 2. This list has also greater diversity in the LLM backbones and sizes. Llama 4 Scout is the only model implementing a mixture of experts (MoE) architecture, which means that only some of the total parameters are activated for each inference step [38]. The inclusion of both models with different sizes of LLM backbones but same VLM finetuning data (e.g. InternVL3 78B and 38B) as well as VLMs with the same LLM backbone and different VLM finetuning data (like Ovis2 16B and InternVL3 14B) allow to draw conclusions for the impact of the LLM, the training data and the vision encoder.

**Table 2.** Overview of VLMs selected for evaluation (*total params, 17B active)

| Model | Params | Image encoder | LLM | OV score |
|---|---|---|---|---|
| InternVL3 78B | 78B | InternViT 6B v2.5 | Qwen 2.5 72B | 79.1 |
| InternVL3 38B | 38B | InternViT 6B v2.5 | Qwen 2.5 32B | 77.8 |
| Ovis2 34B | 34B | AIMv2 1B | Qwen 2.5 32B | 77.0 |
| Qwen2.5VL 72B | 73B | QwenViT | Qwen 2.5 72B | 76.1 |
| InternVL3 14B | 15B | InternViT 300M v2.5 | Qwen 2.5 14B | 75.2 |
| InternVL3 8B | 8B | InternViT 300M v2.5 | Qwen 2.5 7B | 73.6 |
| Ovis2 16B | 16B | AIMv2 Huge | Qwen 2.5 14B | 73.3 |
| Qwen2.5VL 7B | 8B | QwenViT | Qwen 2.5 7B | 70.9 |
| Ovis2 4B | 4.6B | AIMv2 Huge | Qwen 2.5 3B | 69.8 |
| Llama 4 Scout | 109B* | MetaCLIP | Llama 4 | ? |
| Qwen2.5VL 32B | 32B | QwenViT | Qwen 2.5 32B | ? |
| Gemma 3 27B | 27B | SigLIP | Gemma 3 | ? |
| Mistral 3.1 small | 24B | ? | Mistral 3 | ? |

## 5    Evaluation

We ran all models with a size of less than 70B with half precision (FP16) on an Nvidia H100 GPU. Models larger than 70B were run in 4bit AWQ quantization on the same hardware in order to fit them into a single GPU. Although it is common sense that this kind of quantization has only minimal impact on model accuracy, especially for larger models, where it is most relevant [39, 40], we perform some additional tests with a few models in both FP16 and AWQ quantization to assure it is the same in our case.

Since the dataset comprises more than 2,000 questions and the questions are open-ended, we need an automated evaluation that can cope with that. LLM-as-a-judge has proven to be a good candidate to provide better evaluation results for open-ended questions than exact match, ROGUE, BLEU or BERTscore [41], but still has some shortcomings [42, 43]. To compensate for that, the panel of LLM evaluators (PoLL), also known as jury-as-a-judge, uses several independent LLMs to score a model's answer against a known ground truth and uses a voting function to aggregate the independent scores [44]. This could be a majority vote or a simple average.

By scoring each model independently of each other we also avoid common pitfalls like position bias that favors the first answer against later ones in a pairwise comparative assessment [45]. We chose three LLMs as judges that are both very capable and heterogeneous, so that we can expect to have no bias from the same training data or strategy. Mistral 2 large (123B), Llama 3.3 70B and Qwen 2.5 72B come from a European, US and Chinese background and therefore fulfill the criteria. We already had good experiences with using these models in previous studies [19]. Therefore, we decided to keep them instead of exchanging some for more recent models that show slightly better performance in public benchmarks (e.g. Qwen 3 235B-A22B, [46]). We finally used the average of Llama and Mistral since we found Qwen to be overly critical.

One remaining challenge we found in previous studies is that no matter how you tune the prompt for the judge, you always end up with cases where the judge seems too strict or others where it is too generous to grant points to answers that are far from completely correct. An elegant solution we found for this challenge is to identify potential problems on a case-by-case basis and give scoring hints to prevent such problems. This not only improves the correctness of judge scores significantly but also leads to more transparency since these hints are publicly available.

## 6    Results

Overall results show that our benchmark is both challenging and selective. The best model achieves not even 45% accuracy, the worst model scores merely 24%. Results differ between the different domains with best scores in history and religion are around 54%, whereas best scores in music and math are around 31% and 34% respectively.

Table 3.    Results overall and for the first five domains

| Model | Overall | Arts | Biology | Geography | History | Math |
|---|---|---|---|---|---|---|
| InternVL3 8B | 27.92% | 31.82% | 42.02% | 23.51% | 33.71% | 19.32% |
| InternVL3 14B | 32.76% | 37.06% | 40.24% | 29.08% | 39.78% | 27.44% |
| InternVL3 38B | 36.87% | 41.24% | 44.98% | 37.06% | 40.15% | 26.95% |
| InternVL3 78B | 38.15% | 47.03% | 45.86% | 37.65% | 38.45% | 24.51% |
| Ovis2 4B | 23.53% | 31.47% | 32.84% | 26.70% | 31.06% | 9.09% |
| Ovis2 16B | 34.06% | 41.26% | 42.31% | 28.37% | 39.59% | 21.27% |
| Ovis2 34B | 37.12% | 41.09% | 44.08% | 41.84% | 43.18% | 24.35% |
| QwenVL2.5 7B | 32.71% | 41.61% | 36.69% | 30.28% | 38.45% | 22.24% |
| QwenVL2.5 32B | **43.26%** | 41.96% | **48.53%** | 41.04% | **54.36%** | **34.09%** |
| QwenVL2.5 72B | 40.87% | **49.13%** | 47.93% | 44.43% | 42.24% | 29.06% |
| Gemma3 27B | 40.19% | 41.02% | 45.08% | 41.10% | 42.28% | 27.11% |
| Llama4 Scout | 40.29% | 46.88% | 42.22% | 41.85% | 44.35% | 32.63% |
| Mistral small 3.1 | 42.86% | **49.34%** | 46.58% | **50.92%** | 51.97% | 29.06% |

In general, results follow the scaling laws within one family. However, the biggest models do not benefit a lot from the additional size compared to the medium sized ones with 24B-38B parameters (see Table 3). Best scores per category are printed in bold. Second-best values are underlined. QwenVL2.5 72B has a problem in its quantized version that we used and therefore performed worse than the smaller model in some categories. Compared to the MMMU score (the only one that was available for all models, see Table 4) the dominance of InternVL3 and Ovis2 over other model families could not be confirmed. From our benchmark alone, it cannot be concluded whether this is due to German language or due to different kinds of questions, or due to one-sided optimization for well-known public benchmarks. Raw results are published[2].

Table 4.    Results overall and for the remaining four domains (*self-reported)

| Model | Overall | MMMU | Music | Physics | Religion | Sports |
|---|---|---|---|---|---|---|
| InternVL3 8B | 27.92% | 62.2% | 15.67% | 32.48% | 21.85% | 34.09% |
| InternVL3 14B | 32.76% | 64.8% | 24.70% | 35.89% | 25.49% | 34.74% |
| InternVL3 38B | 36.87% | 69.7% | 26.81% | 41.88% | 36.17% | 38.96% |
| InternVL3 78B | 38.15% | **72.2%** | 28.62% | 39.75% | 43.74% | 41.24% |
| Ovis2 4B | 23.53% | 49.0% | 10.24% | 27.78% | 13.84% | 30.20% |
| Ovis2 16B | 34.06% | 60.7% | 23.50% | 38.04% | 31.80% | 37.99% |
| Ovis2 34B | 37.12% | 66.7% | 27.71% | 37.18% | 35.93% | 41.24% |
| QwenVL2.5 7B | 32.71% | 58.0% | 19.88% | 39.32% | 33.74% | 30.20% |
| QwenVL2.5 32B | **43.26%** | 70.0%* | **31.03%** | **47.65%** | 47.33% | **43.84%** |
| QwenVL2.5 72B | 40.87% | 68.2% | 29.22% | 34.83% | 50.49% | 42.21% |
| Gemma3 27B | 40.19% | 64.8% | 30.12% | 45.09% | **52.92%** | 35.07% |
| Llama4 Scout | 40.29% | 69.4%* | 27.72% | 45.30% | 44.91% | 34.10% |
| Mistral small 3.1 | 42.86% | 62.8%* | 25.60% | 46.15% | 48.79% | 32.80% |

## 7      Discussion

We analyzed the wrong model answers in more detail for the best models. Qwen2.5 VL 32B for example scored ~41% in **geography** but had significant problems identifying some of the European islands correctly. In a map with markers numbered from 1-14 and A-H, the model failed to identify Sardinia, Corsica and Cyprus and answered with very wild guesses ("the alps", "Italy" and "Gibraltar"), whereas it was able to correctly name Sicilia and Ireland. Mistral performed even worse in that question.

In the **physics** category, questions that required an explanation were most often answered incorrectly. Both Mistral and Qwen could not explain why the yellow main sequence stars become red giants, when the usual keywords were avoided, so that the clues needed to be read from the image. On the other hand, it was no problem for both models to identify the diagram as a Hertzsprung-Russel diagram. Similarly, both models failed to identify an additional resistor in a parallel circuit as the reason for different light emissions of the light bulbs. Regarding **math**, the models (esp. Mistral) were better at calculating than at counting. For different basic geometric figures, they correctly calculated area and circumference, if the number of corners or sides were counted correctly. However, the octagon was recognized as a decagon and therefore the area was calculated wrongly. Similarly, ten sides were counted, although there were only 9 and again the result got wrong in turn. Both models also failed to identify the hypothenuse of a triangle correctly and therefore could not assign the formula to the respective triangle if the variable names were mixed up compared to the usual $a^2=b^2+c^2$ equation. Similarly, both models struggled to read the coordinate values from a diagram showing geometric shapes with named corner points. They further fail to analyze relatively simple shapes composed of parts of a circle and straight lines in order to calculate their circumference.

---
[2] https://github.com/iisys-hof/vlms-at-school

Models do also have a problem with reverse lookup of colors in a legend, if more than a single entry is concerned. This is a major drawback for answering questions in the **history** section. Furthermore, they also fail to follow arrows backwards and therefore identify the origins of attacks in historic maps. Identifying a city that lies within a named area is not a problem for the model, even if it has to combine information from the historic map with background knowledge of modern city names.

In the **sports** domain, both models fail to combine background knowledge in Basketball or Volleyball with the visual game illustrations that show certain in-game situations. They do not even comprehend that a throw from outside the three-point line is different from a throw inside the line. They also struggle to interpret diagrams.

**Music** is an especially hard challenge for the models. They already struggle to identify simple chords like a C major, or the key given a single cross accidental. Therefore, it is not surprising that they cannot tell which note to change in order to go from one chord to another.

Identifying **art** styles is done correctly for the major, well-known art styles like expressionism and impressionism. For less popular art styles, recognition accuracy drops significantly (e.g., constructivism or existentialism). The painting technique used to create the artwork could only be identified in the rarest of cases.

**Religious** art is identified quite well, if the model is asked for the scene. However, asking further questions based on the identified scene, it seems that the image context does not activate the necessary concepts as a textual context. For example, the creation of Adam and Eve is identified correctly, but the question "what was the woman in the image made of?" is answered incorrectly by Qwen with "from the arm of the man". If the model is asked the questions in sequence retaining the answer from the first question as context for the second question, it is answered correctly. The other way around, sometimes providing a single textual keyword is sufficient to change a completely wrongly identified picture into correctly answered follow-up questions. Qwen is e.g. not able to identify the scene where the people of Israel cross the Red Sea (calling them Huguenots instead) but it correctly identifies the soldiers as belonging to the army of the pharaoh, the man in the middle of the picture as Moses and the water body as the Red Sea. This is especially intriguing, since we carefully chose "water body" (Gewässer) as the keyword, not "sea" which would more clearly hint at Red Sea.

In **biology**, the assignment of technical terms to parts of an illustration using characters, numbers or colors was done with variable accuracy. General terms, that are also used outside the domain experts (e.g., hammer and anvil in the ear) are recognized with higher probability than special terms (e.g., dendrite, semicircular canal). Completely unexpectedly, identifying the owner of the limbs shown in Fig. 1 causes big problems for the models. Only the bird is identified correctly. However, asking the question the other way around, "the limb of which animal is shown under number X" shows, that bird is somewhat the default answer for every question, no matter which number is used for X. Therefore, it seems more like guessing than knowing.

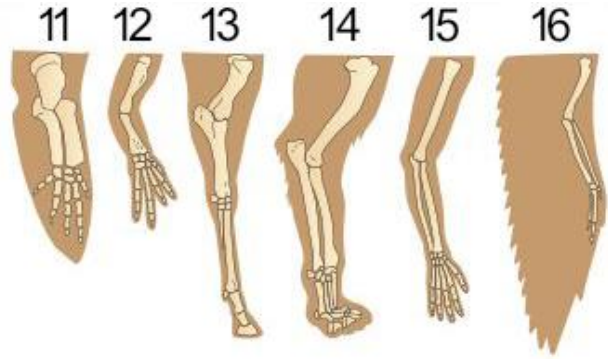

**Fig. 1.** Example question from biology: "Which limb belongs to a whale? State the number!".

### 7.1  Adversarial questions

The answers to the **adversarial questions** were also insightful in the religion domain. The models fell for most of the misleading questions, as expected. Only in very few cases, they correctly answered with "this is not shown in the picture." The bigger models were better. Qwen 2.5 VL 72B took the lead for religion with 9 out of 24 correct answers, whereas most smaller models only gave four correct answers. An extreme example is for the image "The stoning of the adulteress". When asked "Why are the women throwing stones at the man in the middle of the image?" Mistral correctly states that the woman was caught in adultery but then continues with hallucinations that women are throwing stones at Jesus because they think he is going to condemn the woman. Qwen is less obvious wrong. It tells the story correctly as reported in the bible but also fails at stating that there are no women throwing stones at a man (see Fig. 2).

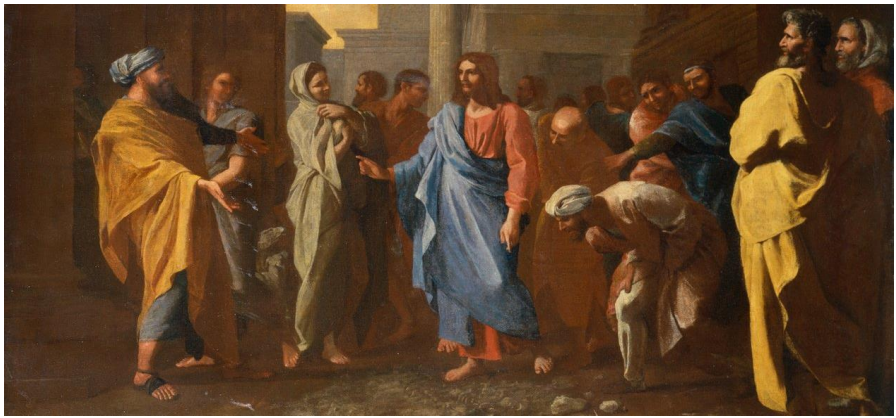

**Fig. 2.** Cropped version of the image "The stoning of the adulteress".

Table 5.   Model performance on adversarial questions overall and in selected domains

| Modell | Overall | Biology | History | Arts | Religion | Music |
|---|---|---|---|---|---|---|
| InternVL3 8B | 10.00% | 11.76% | 13.64% | 11.63% | 12.50% | 7.69% |
| InternVL3 14B | 10.57% | 23.53% | 9.09% | 13.95% | 12.50% | 3.85% |
| InternVL3 38B | 12.83% | 11.76% | 11.36% | 16.28% | 12.50% | 15.38% |
| InternVL3 78B | 12.26% | 17.65% | 2.27% | 16.28% | <u>25.00%</u> | 15.38% |
| Ovis2 4B | 3.02% | 5.88% | 0.00% | 4.65% | 0.00% | 0.00% |
| Ovis2 16B | 13.40% | 11.76% | 4.55% | 20.93% | 16.67% | 15.38% |
| Ovis2 34B | 12.83% | 5.88% | 11.36% | 6.98% | 8.33% | <u>19.23%</u> |
| QwenVL2.5 7B | 15.09% | 11.76% | 11.36% | **37.21%** | 4.17% | 7.69% |
| QwenVL2.5 32B | 29.62% | 47.06% | 40.91% | <u>25.58%</u> | 16.67% | 15.38% |
| QwenVL2.5 72B | 29.62% | 47.06% | 40.91% | <u>25.58%</u> | 16.67% | 15.38% |
| Gemma3 27B | 23.77% | 35.29% | 40.91% | 11.63% | **29.17%** | 11.54% |
| Llama4 Scout | <u>31.32%</u> | <u>52.94%</u> | <u>47.73%</u> | 23.26% | **29.17%** | **26.92%** |
| Mistral small 3.1 | **32.83%** | **64.71%** | **56.82%** | <u>25.58%</u> | 16.67% | 7.69% |

As seen in Table 5 the models' performance in **adversarial questions** is significantly worse than in all questions. An exception is the good scores of a few models like Mistral and Llama 4 in biology and history. In contrast to that, the big InternVL3 models achieve not even a third of the overall accuracy on the adversarial questions. Ovis2 models are also heavily affected. The adversarial questions also show a higher variance across the models. While the ranking within a single domain is mostly consistent with the overall ranking, for the adversarial questions there is a significant difference. Mistral e.g. is the best overall model for adversarial questions, but among the worst for music and, less pronounced, also for religion. On the other hand, the highest scores are also much higher for adversarial questions than overall. The spread between minimum and maximum scores overall are between 9% and 54%, whereas for adversarial questions they are between 0% and 64%.

### 7.2   Comparisons across models

**Comparing models within the same family**, we see that there is a big step in accuracy from the models below 10B parameters to the medium sized models (14B - 38B), but only a small increase for the next scaling step from 32B or 38B to over 70B parameters. Scaling the vision encoder as done by InternVL3 from 300M (14B model) to 6B (38B model) does not result in a significant increase in accuracy. Compared to the Ovis2 model family with the same LLM backend and only a moderate increase in size of the vision encoder (300M => 1B), we cannot observe a benefit of the bigger vision encoder.

**Comparing models** of similar size **across model families**, we see that both Mistral 3.1 small 24B as well as Qwen2.5 VL 32B perform exceptionally well. Gemma3 is also well above the scaling trendline of InternVL3 and Ovis2 (see Fig. 3).

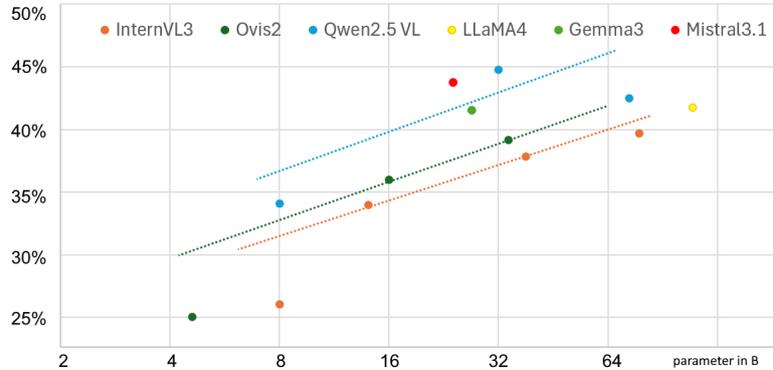

**Fig. 3.** Model size (log$_2$) vs. overall accuracy for three model families.

Comparing the models' **MMMU score against their performance in our dataset** (see Fig. 4), we see that Qwen 2.5 VL shows roughly a linear relation, whereas InternVL3 significantly underperforms on our dataset compared to MMMU. Ovis2 is within the linear trend for 4B and 16B but underperforms for 34B. Gemma3 and Llama3 are slightly above and below the trendline respectively. Only Mistral significantly deviates positively from the linear trend.

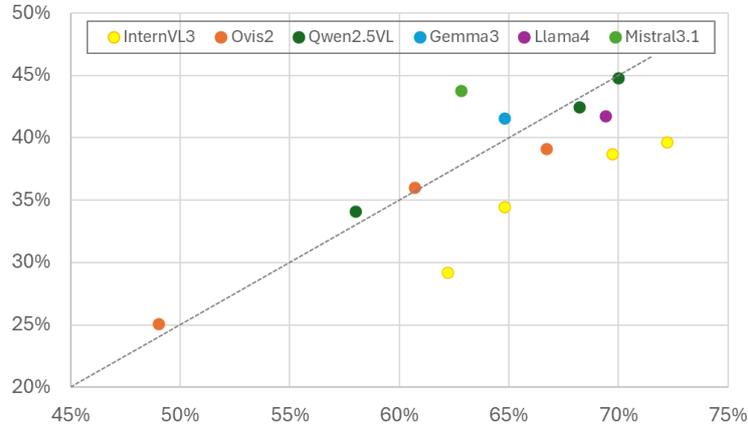

**Fig. 4.** MMMU score (x-axis) vs. VLM@school score (y-axis).

## 8   Limitations

We ran the biggest models (>70B parameters) only in a quantized version to fit them into a single H100 GPU. Comparing the quantized models' results with those of the models tested with half precision (FP16) confirms the hypothesis, that quantization leads to worse performance, but the accuracy degradation stays within acceptable bounds, especially for bigger models, where it is more important to use a quantized version instead of an FP16 version. Where 3% absolute can be seen as a significant quality drop for the 8B parameter model (a difference of nearly 11% relative), the 14B and 38B models show a small degradation of 0.45% and 0.83% respectively. Therefore, degradation of the biggest models that were run only in a quantized version are expected to stay within the same bounds of 1% absolute in the worst case and <0.5% on average. So no fundamental changes compared to the reported scores are expected (see Table 6).

Table 6. Comparison of half-precision vs. AWQ scores for InternVL3 models

| Model | AWQ score | FP16 score | Delta |
|---|---|---|---|
| InternVL3 8B | 26.10% | 29.24% | 3.14% |
| InternVL3 14B | 34.00% | 34.45% | 0.45% |
| InternVL3 38B | 37.88% | 38.71% | 0.83% |
| InternVL3 78B | 39.70% | - | - |

Despite our best effort to keep questions subject specific, there is some overlap in types of questions across domains. We have e.g. map-related questions not only in geography, but also in history. Interpretation of diagrams and tables is required in physics and biology.

Although we find human baselines very important, we were not able to find enough people to get a statistically significant result for answering all 2000 questions. Since many questions require thinking, answering all questions would take several hours. We don't think that it makes sense to hire crowd workers via Amazon Mechanical Turk for this task. We expect, however, average humans to also not reach >90% because although the necessary knowledge for each category is only middle school level, it is quite diverse and most people are only really good in a few selected categories, but not across the board. If we would have answers from a school class (10[th] grade) we would expect best humans per category to reach 90% and more, but only about 60% in general.

Even open-ended questions can have a limited number of meaningful answers. We did our best to minimize chances of guessing the right answer, but there are still a few questions where we cannot exclude this. We did not systematically measure guessing probabilities except the LLM-only answers discussed in section 2 that lead to 8.5% correct answers for the best of three tested models and 5% for the worst.

We did not test any commercial API-only models like GPT4o, Gemini 2.5 Pro or Claude 4 Opus which obtain high ranks in public benchmarks. We have no subscription and didn't want to pay money to test these models with all 2000 questions because we feel that the commercial providers should rather pay us to test their models for them. We did, however, do a few random samples and found the models to perform well, but also struggle with several types of questions. So, we expect them to take a lead in our ranking, but rather in the range of 50-60% and not above 80% accuracy.

We had the suspicion that the good results in identifying religious scenes on images could be due to the use of well-known paintings of famous artists that were part of the training data. The association of the name of the painting alone would be enough to provide enough hints for an easy identification. Therefore, we tested with a dozen AI generated images (self-generated with Flux) to see whether results are worse for those images that were surely not included in the training data. Surprisingly, these images were identified with similar accuracy. Our suspicions could not be confirmed.

We had planned to include chemistry as a tenth domain in our dataset. However, it turned out to be hard to find enough images and phrase questions without enough hints to prevent LLMs from answering the question without the image. We therefore skipped that domain completely.

## 9      Conclusion and Future Work

We presented a new challenging dataset with over 2,000 question-answer pairs on 486 images to evaluate Vision Language Models (VLMs) on nine different school subjects in German language. We showed that it doesn't need complex questions at university or even PhD level to highlight the limitations of existing AI solutions. Middle school grade questions are challenging enough, if the models have to combine visual clues with background knowledge. We further showed that adversarial questions that ask for things not seen in the provided image are mostly answered incorrectly. AI models still haven't learned to say "no", if the context of the question is semantically close to the image.

Our analysis reveals that results on public benchmarks do not necessarily reflect a model's accuracy in other use cases, even if they are closely related. The InternVL3 model family dominates the Open VLM leaderboard and equally the scores for MMMU, a benchmark with high similarity to our own dataset. However, InternVL3 models do not perform exceptionally well on our dataset and are outperformed by Mistral 3.1, Qwen 2.5 VL 32B and Gemma 3 27B. Especially for adversarial questions the gap of the InternVL3 models to the rest of the competition is obvious, where even the biggest 78B model achieves roughly a third of the best competitors, although they are much smaller.

In the future, we should prepare a version of the dataset with English questions and answers to find out whether the German language is responsible for the low accuracy of model answers. Our current hypothesis is that the rigorous testing for not including questions that can be answered from the text alone are the reason. We carefully avoided keywords that provide too many hints and used larger numbers (>5) and characters from the second half of the alphabet instead of A-D for labeling image parts to minimize chances of correct guessing. It seems that visual clues are not equally efficient to trigger a certain context that helps to find the correct answers as textual hints are. We hypothesize that chain-of-thought or reasoning approaches, that first translate the models' "thoughts" about the image into explicit text before answering the question would boost performance substantially [47]. A few tests with Kimi VL A3B, a 16B MoE model [48] and to the best of the authors knowledge the only open-weight VLM available with reasoning finetuning, showed a weak performance, so that we did not include this model into the final comparison. The runtime for answers was prohibitively long and we did not expect any further insight.

Another suitable approach to better understand the models' capabilities would be to use the images and questions of MMMU-Pro as a basis, translate them to German and convert them to open-ended questions to compare results of this benchmark and ours.

# References

1. Russakovsky, O., Deng, J., Su, H., Krause, J., Satheesh, S., Ma, S., Huang, Z., Karpathy, A., Khosla, A., Bernstein, M., Berg, A.C., Fei-Fei, L.: ImageNet Large Scale Visual Recognition Challenge. Int. J. Comput. Vis. 115, 211–252 (2015).
2. Panayotov, V., Chen, G., Povey, D., Khudanpur, S.: Librispeech: an asr corpus based on public domain audio books. In: 2015 IEEE international conference on acoustics, speech and signal processing (ICASSP). pp. 5206–5210. IEEE (2015).
3. Wang, Y., Ma, X., Zhang, G., Ni, Y., Chandra, A., Guo, S., Ren, W., Arulraj, A., He, X., Jiang, Z.: MMLU-Pro: A More Robust and Challenging Multi-Task Language Understanding Benchmark. Adv. Neural Inf. Process. Syst. 37, 95266–95290 (2025).
4. Eriksson, M., Purificato, E., Noroozian, A., Vinagre, J., Chaslot, G., Gomez, E., Fernandez-Llorca, D.: Can We Trust AI Benchmarks? An Interdisciplinary Review of Current Issues in AI Evaluation, https://arxiv.org/abs/2502.06559v1, last accessed 2025/05/26.
5. Raji, I.D., Bender, E.M., Paullada, A., Denton, E., Hanna, A.: AI and the Everything in the Whole Wide World Benchmark, http://arxiv.org/abs/2111.15366, (2021).
6. Weij, T. van der, Hofstätter, F., Jaffe, O., Brown, S.F., Ward, F.R.: AI Sandbagging: Language Models can Strategically Underperform on Evaluations, http://arxiv.org/abs/2406.07358, (2025). https://doi.org/10.48550/arXiv.2406.07358.
7. Oakden-Rayner, L., Dunnmon, J., Carneiro, G., Re, C.: Hidden stratification causes clinically meaningful failures in machine learning for medical imaging. In: Proceedings of the ACM Conference on Health, Inference, and Learning. pp. 151–159. ACM, Toronto Ontario Canada (2020). https://doi.org/10.1145/3368555.3384468.
8. Ethayarajh, K., Jurafsky, D.: Utility is in the Eye of the User: A Critique of NLP Leaderboards, http://arxiv.org/abs/2009.13888, (2021).
9. Reuel-Lamparth, A., Hardy, A., Smith, C., Lamparth, M., Hardy, M., Kochenderfer, M.J.: BetterBench: Assessing AI Benchmarks, Uncovering Issues, and Establishing Best Practices. Adv. Neural Inf. Process. Syst. 37, 21763–21813 (2024).
10. Dehghani, M., Tay, Y., Gritsenko, A.A., Zhao, Z., Houlsby, N., Diaz, F., Metzler, D., Vinyals, O.: The Benchmark Lottery, http://arxiv.org/abs/2107.07002, (2021).
11. Kang, E.B.: Ground truth tracings (GTT): On the epistemic limits of machine learning. Big Data Soc. 10, 20539517221146122 (2023).
12. Peinl, R., Wirth, J.: Evaluation of Medium-Sized Language Models in German and English Language. Int. J. Nat. Lang. Comput. 13, 01–18 (2024).
13. Mizrahi, M., Kaplan, G., Malkin, D., Dror, R., Shahaf, D., Stanovsky, G.: State of what art? A call for multi-prompt LLM evaluation. Trans. Assoc. Comput. Linguist. 12, 933–949 (2024).
14. Rauh, M., Marchal, N., Manzini, A., Hendricks, L.A., Comanescu, R., Akbulut, C., Stepleton, T., Mateos-Garcia, J., Bergman, S., Kay, J.: Gaps in the Safety Evaluation of Generative AI. In: AAAI/ACM Conf. on AI, Ethics, and Society. pp. 1200–1217 (2024).
15. Yue, X., Ni, Y., Zhang, K., Zheng, T., Liu, R., Zhang, G., Stevens, S., Jiang, D., Ren, W., Sun, Y.: MMMU: A massive multi-discipline multimodal understanding and reasoning benchmark for expert AGI. In: IEEE/CVF Conference on Computer Vision and Pattern Recognition. pp. 9556–9567 (2024).
16. Duan, H., Yang, J., Qiao, Y., Fang, X., Chen, L., Liu, Y., Dong, X., Zang, Y., Zhang, P., Wang, J., Lin, D., Chen, K.: VLMEvalKit: An Open-Source ToolKit for Evaluating Large Multi-Modality Models. In: 32nd ACM International Conference on Multimedia. pp. 11198–11201. ACM, Melbourne VIC Australia (2024).
17. Announcing ARC-AGI-2 and ARC Prize 2025, https://arcprize.org/blog/announcing-arc-agi-2-and-arc-prize-2025, last accessed 2025/05/26.
18. Phan, L., Gatti, A., Han, Z., Li, N., Hu, J., Zhang, H., Zhang, C.B.C., Shaaban, M., Ling, J., Shi, S.: Humanity's last exam. ArXiv Prepr. ArXiv250114249. (2025).

19. Peinl, R., Tischler, V.: Benchmarking Vision Language Models on German Factual Data. Presented at the 21st Int. Conf. on Artificial Intelligence Applications and Innovations , Limassol, Cyprus June 26 (2025).
20. Gemini 2.5: Our most intelligent AI model, https://blog.google/technology/google-deepmind/gemini-model-thinking-updates-march-2025/, last accessed 2025/05/26.
21. The Llama 4 herd: The beginning of a new era of natively multimodal AI innovation, https://ai.meta.com/blog/llama-4-multimodal-intelligence/, last accessed 2025/05/26.
22. Introducing Claude 4, https://www.anthropic.com/news/claude-4, last accessed 2025/05/26.
23. Yue, X., Zheng, T., Ni, Y., Wang, Y., Zhang, K., Tong, S., Sun, Y., Yu, B., Zhang, G., Sun, H., Su, Y., Chen, W., Neubig, G.: MMMU-Pro: A More Robust Multi-discipline Multimodal Understanding Benchmark, http://arxiv.org/abs/2409.02813, (2025).
24. Masry, A., Long, D.X., Tan, J.Q., Joty, S., Hoque, E.: ChartQA: A Benchmark for Question Answering about Charts with Visual and Logical Reasoning, http://arxiv.org/abs/2203.10244, (2022). https://doi.org/10.48550/arXiv.2203.10244.
25. Kembhavi, A., Salvato, M., Kolve, E., Seo, M., Hajishirzi, H., Farhadi, A.: A Diagram is Worth a Dozen Images. In: Leibe, B., Matas, J., Sebe, N., and Welling, M. (eds.) Computer Vision – ECCV 2016. pp. 235–251. Springer, Cham (2016).
26. Yu, W., Yang, Z., Li, L., Wang, J., Lin, K., Liu, Z., Wang, X., Wang, L.: MM-Vet: Evaluating Large Multimodal Models for Integrated Capabilities, http://arxiv.org/abs/2308.02490, (2024). https://doi.org/10.48550/arXiv.2308.02490.
27. Guo, Z., Zhang, R., Chen, H., Gao, J., Jiang, D., Wang, J., Heng, P.-A.: SciVerse: Unveiling the Knowledge Comprehension and Visual Reasoning of LMMs on Multi-modal Scientific Problems, http://arxiv.org/abs/2503.10627, (2025).
28. Mathew, M., Karatzas, D., Jawahar, C.V.: DocVQA: A dataset for VQA on document images. In: IEEE/CVF Winter Conf. on Applications of Computer Vision. pp. 2200–2209 (2021).
29. Liu, Y., Li, Z., Huang, M., Yang, B., Yu, W., Li, C., Yin, X.-C., Liu, C.-L., Jin, L., Bai, X.: OCRBench: on the hidden mystery of OCR in large multimodal models. Sci. China Inf. Sci. 67, 220102 (2024). https://doi.org/10.1007/s11432-024-4235-6.
30. Xing, S., Sun, Z., Xie, S., Chen, K., Huang, Y., Wang, Y., Li, J., Song, D., Tu, Z.: Can Large Vision Language Models Read Maps Like a Human?, http://arxiv.org/abs/2503.14607, (2025). https://doi.org/10.48550/arXiv.2503.14607.
31. Das, R.J., Hristov, S.E., Li, H., Dimitrov, D.I., Koychev, I., Nakov, P.: EXAMS-V: A Multi-Discipline Multilingual Multimodal Exam Benchmark for Evaluating Vision Language Models, http://arxiv.org/abs/2403.10378, (2024).
32. Sun, H.-L., Zhou, D.-W., Li, Y., Lu, S., Yi, C., Chen, Q.-G., Xu, Z., Luo, W., Zhang, K., Zhan, D.-C., Ye, H.-J.: Parrot: Multilingual Visual Instruction Tuning, http://arxiv.org/abs/2406.02539, (2024). https://doi.org/10.48550/arXiv.2406.02539.
33. Li, L., Yin, Y., Li, S., Chen, L., Wang, P., Ren, S., Li, M., Yang, Y., Xu, J., Sun, X.: M^3 IT: A Large-Scale Dataset towards Multi-Modal Multilingual Instruction Tuning. ArXiv Prepr. ArXiv230604387. (2023).
34. Pfeiffer, J., Geigle, G., Kamath, A., Steitz, J.-M.O., Roth, S., Vulić, I., Gurevych, I.: xGQA: Cross-Lingual Visual Question Answering, http://arxiv.org/abs/2109.06082, (2022).
35. Park, C., Lee, K., Lim, H., Kim, J., Park, J., Heo, Y.-J., Chang, D.-S., Choo, J.: Translation Deserves Better: Analyzing Translation Artifacts in Cross-lingual Visual Question Answering, http://arxiv.org/abs/2406.02331, (2024).
36. Deng, Y., Zhao, Y., Li, M., Ng, S.-K., Chua, T.-S.: Don't Just Say "I don't know"! Self-aligning Large Language Models for Responding to Unknown Questions with Explanations, http://arxiv.org/abs/2402.15062, (2024).
37. Brahman, F., Kumar, S., Balachandran, V., Dasigi, P., Pyatkin, V., Ravichander, A., Wiegreffe, S., Dziri, N., Chandu, K., Hessel, J.: The art of saying no: Contextual noncompliance in language models. Adv. Neural Inf. Process. Syst. 37, 49706–49748 (2024).
38. Lin, B., Tang, Z., Ye, Y., Cui, J., Zhu, B., Jin, P., Huang, J., Zhang, J., Ning, M., Yuan, L.: MoE-LLaVA: Mixture of Experts for Large Vision-Language Models, (2024).
39. Lin, J., Tang, J., Tang, H., Yang, S., Dang, X., Han, S.: AWQ: Activation-aware Weight Quantization for LLM Compression and Acceleration, (2023).
40. Wang, C., Wang, Z., Xu, X., Tang, Y., Zhou, J., Lu, J.: Q-VLM: Post-training Quantization for Large Vision-Language Models. In: 38 Annual Conference on Neural Information Processing Systems (2024).
41. Zhu, L., Wang, X., Wang, X.: JudgeLM: Fine-tuned Large Language Models are Scalable Judges, (2023).
42. Huang, H., Qu, Y., Bu, X., Zhou, H., Liu, J., Yang, M., Xu, B., Zhao, T.: An Empirical Study of LLM-as-a-Judge for LLM Evaluation: Fine-tuned Judge Model is not a General Substitute for GPT-4, http://arxiv.org/abs/2403.02839, (2024).
43. Raina, V., Liusie, A., Gales, M.: Is LLM-as-a-Judge Robust? Investigating Universal Adversarial Attacks on Zero-shot LLM Assessment, (2024).
44. Verga, P., Hofstatter, S., Althammer, S., Su, Y., Piktus, A., Arkhangorodsky, A., Xu, M., White, N., Lewis, P.: Replacing Judges with Juries: Evaluating LLM Generations with a Panel of Diverse Models, http://arxiv.org/abs/2404.18796, (2024).
45. Shi, L., Ma, C., Liang, W., Ma, W., Vosoughi, S.: Judging the judges: A systematic investigation of position bias in pairwise comparative assessments by llms. ArXiv Prepr. ArXiv240607791. (2024).
46. Alibaba: Qwen3: Think Deeper, Act Faster, https://qwenlm.github.io/blog/qwen3/, last accessed 2025/05/27.
47. Xu, G., Jin, P., Li, H., Song, Y., Sun, L., Yuan, L.: LLaVA-CoT: Let Vision Language Models Reason Step-by-Step, http://arxiv.org/abs/2411.10440, (2025).
48. Team Kimi: Kimi-VL Technical Report, http://arxiv.org/abs/2504.07491, (2025). https://doi.org/10.48550/arXiv.2504.07491.